\documentclass[11pt]{article}
\newlength{\defbaselineskip}
\usepackage{hyperref}
\usepackage{url}

\usepackage{algorithm}
\usepackage{algpseudocode}
\usepackage{framed}
\usepackage{amssymb}
\usepackage{amsfonts}
\usepackage{amsmath}
\usepackage{graphicx}
\usepackage{url}
\usepackage{rotating}
\usepackage{multirow}
\usepackage{color}
\usepackage{xcolor}
\usepackage{paralist}

\usepackage{subfigure}

\usepackage{longtable} 
\usepackage{makecell}

\usepackage[multiple]{footmisc}
\usepackage[section]{placeins}

\usepackage{hyperref}
\hypersetup{
     colorlinks   = true,
     linkcolor    = blue,
     citecolor    = green
}

\usepackage[normalem]{ulem}

\usepackage{enumitem}
\setitemize{noitemsep,topsep=0pt,parsep=0pt,partopsep=0pt,leftmargin=*}

\begin{document}

\title{%
Traditional and Heavy-Tailed Self Regularization in Neural Network Models%
\footnote{This is a short version of a longer and more detailed paper~\cite{MM18_TR}.  See the long version for many more results and a much more detailed exposition.  Also, subsequent to the posting of the long version of this paper, this theory has been used to construct a Universal capacity control metric that predicts trends in test accuracies for very large pre-trained deep neural networks~\cite{MM19a_TR}. }
}

\author{%
Charles H. Martin\thanks{Calculation Consulting, 8 Locksley Ave, 6B, San Francisco, CA 94122, \texttt{charles@CalculationConsulting.com}.} 
\and 
Michael W. Mahoney\thanks{ICSI and Department of Statistics, University of California at Berkeley, Berkeley, CA 94720, \texttt{mmahoney@stat.berkeley.edu}.}
}

\date{}
\maketitle

%\begin{abstract}
%The abstract paragraph should be indented 1/2~inch (3~picas) on both left and
%right-hand margins. Use 10~point type, with a vertical spacing of 11~points.
%The word \textsc{Abstract} must be centered, in small caps, and in point size 12. Two
%line spaces precede the abstract. The abstract must be limited to one
%paragraph.
%\end{abstract}

\begin{abstract}
\noindent
Random Matrix Theory (RMT) is applied to analyze the weight matrices of Deep Neural Networks (DNNs), including both production quality, pre-trained models such as AlexNet and Inception, and smaller models trained from scratch, such as LeNet5 and a miniature-AlexNet.
Empirical and theoretical results clearly indicate that
the empirical spectral density (ESD) of DNN layer matrices displays signatures of traditionally-regularized statistical models, even in the absence of exogenously specifying traditional forms of 
regularization, such as Dropout or Weight Norm constraints.
Building on 
recent results in RMT, most notably its extension to Universality classes of Heavy-Tailed matrices, 
we develop a theory to identify \emph{5+1 Phases of Training}, corresponding to increasing amounts of \emph{Implicit Self-Regularization}.
For smaller and/or older DNNs, this Implicit Self-Regularization is like traditional Tikhonov regularization, in that there is a ``size scale'' separating signal from noise.
For state-of-the-art DNNs, however, we identify a novel form of \emph{Heavy-Tailed Self-Regularization}, similar to the self-organization seen in the statistical physics of disordered systems. 
This implicit Self-Regularization can depend strongly on the many knobs of the training process.
By exploiting the generalization gap phenomena,
we demonstrate that we can cause a small model to exhibit all 5+1 phases of training simply by changing the batch size.
\end{abstract}

\section{Introduction}
\label{sxn:intro}

The inability of optimization and learning theory to explain and predict the properties of NNs is not a new phenomenon.
From the earliest days of DNNs, it was suspected that VC theory did not apply to these systems~\cite{VLL94}. 
It was originally assumed that local minima in the energy/loss surface were responsible for the inability of VC theory to describe NNs~\cite{VLL94}, and that the mechanism for this was that getting trapped in local minima during training limited the number of possible functions realizable by the network.
However, it was very soon realized that the presence of local minima in the energy function was \emph{not} a problem in practice~\cite{LBBH98,orangeBook}.
Thus, another reason for the inapplicability of VC theory was needed.
At the time, there did exist other theories of generalization based on statistical mechanics~\cite{SST92,WRB93,DKST96,EB01_BOOK}, but for various technical and nontechnical reasons these fell out of favor in the ML/NN communities.
Instead, VC theory and related techniques continued to remain popular, in spite of their obvious~problems.

More recently,
theoretical results of Choromanska et al.~\cite{CHMAx14_TR} (which are related to~\cite{SST92,WRB93,DKST96,EB01_BOOK}) suggested that the Energy/optimization Landscape of modern DNNs resembles the Energy Landscape of a zero-temperature \emph{Gaussian} Spin Glass; and 
empirical results of Zhang et al.~\cite{Understanding16_TR} have again pointed out that VC theory does not describe the properties of DNNs.
Martin and Mahoney then suggested that the Spin Glass analogy may be useful to understand severe overtraining versus the inability to overtrain in modern DNNs~\cite{MM17_TR}.

We should note that
it is not even clear how to define DNN regularization. 
The challenge in applying these well-known ideas to DNNs is that DNNs have \emph{many} adjustable ``knobs and switches,'' independent of the Energy Landscape itself, most of which can affect training accuracy, in addition to \emph{many} model parameters.
Indeed, nearly anything that improves generalization is called regularization~\cite{KGC17_TR}.  
Evaluating and comparing these methods is challenging, in part since there are so many, and in part since they are often constrained by systems or other not-traditionally-ML considerations.

Motivated by this situation, we are interested here in two related questions.
\begin{itemize}
\item
\textbf{Theoretical Question.}
Why is regularization in deep learning seemingly quite different than regularization in other areas on ML; and 
what is the right theoretical framework with which to investigate
regularization for DNNs?
\item
\textbf{Practical Question.}
How can one control and adjust, in a theoretically-principled way, the many knobs and switches that exist in modern DNN systems, e.g., to train these models efficiently and effectively, to monitor their effects on the global Energy Landscape, etc.?
\end{itemize}

\noindent
That is, we seek a \emph{Practical Theory of Deep Learning}, one that is prescriptive and not just descriptive.
This theory would provide useful tools for practitioners wanting to know \emph{How} to characterize and control the Energy Landscape to engineer larger and betters DNNs; and it would also provide theoretical answers to broad open questions as \emph{Why} Deep Learning even~works.

\paragraph{Main Empirical Results.}
Our main empirical results consist in evaluating empirically the ESDs (and related RMT-based statistics) for weight matrices for a suite of DNN models, thereby probing the Energy Landscapes of these DNNs.
For older and/or smaller models, these results are consistent with implicit \emph{Self-Regularization} that is Tikhonov-like; and for modern state-of-the-art models, these results suggest novel forms of \emph{Heavy-Tailed Self-Regularization}.

\begin{itemize}

\item 
\textbf{Self-Regularization in old/small models.}
The ESDs of older/smaller DNN models (like LeNet5 and a toy MLP3 model) exhibit weak \emph{Self-Regularization}, well-modeled by a perturbative variant of MP theory, the Spiked-Covariance model. 
Here, a small number of eigenvalues pull out from the random bulk, and thus the MP Soft Rank and Stable Rank both decrease.
This weak form of \emph{Self-Regularization} is like Tikhonov regularization, in that there is a ``size scale'' that cleanly separates ``signal'' from ``noise,'' but it is different than explicit Tikhonov regularization in that it arises implicitly due to the DNN training process itself.

\item 
\textbf{Heavy-Tailed Self-Regularization.}
The ESDs of larger, modern DNN models (including AlexNet and Inception and nearly every other large-scale model we have examined) deviate strongly from the common Gaussian-based MP model.
Instead, they appear to lie in one of the very different Universality classes of Heavy-Tailed random matrix models.
We call this \emph{Heavy-Tailed Self-Regularization}.
The ESD appears 
Heavy-Tailed, but with finite support.
In this case, there is \emph{not} a ``size scale'' (even in the theory) that cleanly separates ``signal'' from~``noise.'' 

\end{itemize}

\paragraph{Main Theoretical Results.}
Our main theoretical results consist in an operational theory for DNN Self-Regularization.
Our theory uses ideas from RMT---both vanilla MP-based RMT as well as extensions to other Universality classes based on Heavy-Tailed distributions---to provide a visual taxonomy for \emph{$5+1$ Phases of Training}, corresponding to increasing amounts of Self-Regularization.

\begin{itemize}

\item 
\textbf{Modeling Noise and Signal.}
We assume that a weight matrix $\mathbf{W}$ can be modeled as
$
\mathbf{W}\simeq\mathbf{W}^{rand}+\Delta^{sig}   ,
$
where $\mathbf{W}^{rand}$ is ``noise'' and where $\Delta^{sig}$ is ``signal.''
For small to medium sized signal, $\mathbf{W}$ is well-approximated by an MP distribution---with elements drawn from the Gaussian Universality class---perhaps after removing a few eigenvectors.
For large and strongly-correlated signal, $\mathbf{W}^{rand}$ gets progressively smaller, but we can model the non-random strongly-correlated signal $\Delta^{sig}$ by a Heavy-Tailed random matrix, i.e., a random matrix with elements drawn from a Heavy-Tailed (rather than Gaussian) Universality class.

\item 
\textbf{5+1 Phases of Regularization.}
Based on this,
we construct a practical, visual taxonomy for 5+1 Phases of Training.
Each phase is characterized by stronger, visually distinct signatures in the ESD of DNN weight matrices, and successive phases correspond to decreasing MP Soft Rank and increasing amounts of \emph{Self-Regularization}. 
The 5+1 phases are: \textsc{Random-like}, \textsc{Bleeding-out}, \textsc{Bulk+Spikes}, \textsc{Bulk-decay}, \textsc{Heavy-Tailed}, and \textsc{Rank-collapse}.

\end{itemize}

\noindent
Based on these results, we speculate that all well optimized, large DNNs will display \emph{Heavy-Tailed Self-Regularization} in their weight matrices.  %, even when combined in different ways. 

\paragraph{Evaluating the Theory.}
We provide a detailed evaluation of our theory using a smaller MiniAlexNew model that we can train and retrain.

\begin{itemize}

\item
\textbf{Effect of Explicit Regularization.}
We analyze ESDs of MiniAlexNet by removing all explicit regularization (Dropout, Weight Norm constraints, Batch Normalization, etc.) and characterizing how the ESD of weight matrices behave during and at the end of Backprop training, as we systematically add back in different forms of explicit regularization.

\item
\textbf{Exhibiting the 5+1 Phases.}
We demonstrate that we can exhibit all 5+1 phases by appropriate modification of the various knobs of the training process. 
In particular, by decreasing the batch size from 500 to 2, we can make the ESDs of the fully-connected layers of MiniAlexNet vary continuously from \textsc{Random-like} to \textsc{Heavy-Tailed}, while increasing generalization accuracy along the way.
These results illustrate the \emph{Generalization Gap} pheneomena~\cite{HHS17_TR, KMNST16_TR, one_hour17_TR}, and they explain that pheneomena as being caused by the implicit Self-Regularization associated with models trained with smaller and smaller batch sizes.

\end{itemize}

\section{Basic Random Matrix Theory (RMT)}
\label{sxb:review-RMT}

In this section, we summarize results from RMT that we use.
Several overviews of RMT are available~\cite{ TV04, ER05, Kar05_recent, TW09, bouchaud2009, EW13, PA14, bun2017}.
Here, we will describe a more general form of RMT.

\subsection{Marchenko-Pastur (MP) theory for rectangular matrices} 
\label{sxb:review-RMT-MP}

MP theory considers the density of singular values $\rho(\nu_{i})$ of random rectangular matrices $\mathbf{W}$.
This is equivalent to considering the density of eigenvalues $\rho(\lambda_{i})$, i.e., the ESD, of matrices of the form $\mathbf{X}=\mathbf{W}^{T}\mathbf{W}$.
MP theory then makes strong statements about such quantities as the shape of the distribution in the infinite limit, it's bounds, expected finite-size effects, such as fluctuations near the edge, and rates of convergence.

To apply RMT, we need only specify the number of rows and columns of $\mathbf{W}$ and assume that the elements $W_{i,j}$ are drawn from a 
distribution that is a member of a certain \emph{Universality class} (there are different results for different Universality classes).
RMT then describes properties of the ESD, even at finite size; and one can compare perdictions of RMT with empirical results.  
Most well-known 
is the Universality class of Gaussian distributions.
This leads to the basic or vanilla MP theory, which we describe in this section.
More esoteric---but ultimately more useful for us---are Universality classes of Heavy-Tailed distributions.
In Section~\ref{sxb:review-RMT-extensions}, we describe this important variant.

\noindent\textbf{Gaussian Universality class.}
We start
by modeling $\mathbf{W}$ as an $N\times M$ random matrix, with elements 
from a Gaussian distribution, such that: 
$ %%$$
W_{ij} \sim N(0,\sigma_{mp}^2)  .
$ %%$$
Then, MP theory states that the ESD of the correlation matrix, $\mathbf{X}=\mathbf{W}^{T}\mathbf{W}$, has the limiting density given by the MP distribution $\rho(\lambda)$:  
\begin{eqnarray}
\nonumber
\rho_N(\lambda) 
\label{eqn:mp_distribution}
   &\xrightarrow[Q\;\text{fixed}]{N \rightarrow \infty}& 
      \left\{ \begin{array}{ll}
                  \dfrac{Q}{2\pi\sigma_{mp}^{2}}\dfrac{\sqrt{(\lambda^{+}-\lambda)(\lambda-\lambda^{-})}}{\lambda}   & \mbox{if $\lambda\in[\lambda^{-},\lambda^{+}]$} \\
                 0 & \mbox{otherwise}   .
              \end{array}
      \right.  
\end{eqnarray}
Here, $\sigma^2_{mp}$ is the element-wise variance of the original matrix, $Q=N/M\ge1$ is the aspect ratio of the matrix, and the minimum and maximum eigenvalues, $\lambda^{\pm}$, are given by 
\begin{equation}
\lambda^{\pm}=\sigma_{mp}^{2}\left(1\pm\dfrac{1}{\sqrt Q}\right)^{2}  .
\label{eqn:lambda_pm}
\end{equation}

\noindent\textbf{Finite-size Fluctuations at the MP Edge.} 
In the infinite limit, all fluctuations in $\rho_N(\lambda)$ concentrate very sharply at the MP edge, $\lambda^{\pm}$, and the distribution of the maximum eigenvalues $\rho_{\infty}(\lambda_{max})$ is governed by the TW Law. 
Even for a single finite-sized matrix, however, MP theory states the upper edge of $\rho(\lambda)$ is very sharp; and even when the MP Law is violated, the TW Law, with finite-size corrections, works very well at describing the edge statistics.
When these laws are violated, this is very strong evidence for the onset of more regular non-random structure in the DNN weight matrices, which we will interpret as evidence of \emph{Self-Regularization}.

\subsection{Heavy-Tailed extensions of MP theory}
\label{sxb:review-RMT-extensions}

\begin{table}[t]
\small
\begin{center}
\begin{tabular}{|p{1in}|c|c|c|c|c|c|c|}
\hline
 
 & \makecell{Generative Model \\ w/ elements from \\  Universality class }
 & \makecell{Finite-$N$\\ Global shape \\  $\rho_N(\lambda)$} 
 & \makecell{Limiting\\ Global shape \\ $\rho(\lambda),\;N\rightarrow\infty$ }
 & \makecell{Bulk edge \\ Local stats \\  $\lambda \approx \lambda^{+}$  }
 & \makecell{ (far) Tail \\  Local stats \\ $\lambda \approx \lambda_{max}$  } \\
\hline
\hline
%\hline
\makecell{Basic MP}
 & \makecell{ Gaussian  }
 & \makecell{ MP, i.e., \\ Eqn.~(\ref{eqn:mp_distribution}) }
 & MP  
 & \makecell{ TW }
 & No tail. \\
\hline
\makecell{Spiked-\\Covariance}
 & \makecell{ Gaussian, \\ + low-rank \\ perturbations }
 & \makecell{ MP + \\ Gaussian \\ spikes }
 & MP  
 & \makecell{  TW }
 & \makecell{  Gaussian } \\
\hline
\makecell{Heavy tail, \\ $4 < \mu $}
 & \makecell{ (Weakly) \\ Heavy-Tailed  }
 & \makecell{ MP + \\ PL tail }
 & MP  
 & \makecell{  Heavy-Tailed$^{*}$  }
 & \makecell{  Heavy-Tailed$^{*}$  } \\
\hline
\makecell{Heavy tail, \\ $2 < \mu < 4$}
 & \makecell{ (Moderately) \\ Heavy-Tailed \\ (or ``fat tailed'') }
 & \makecell{ PL$^{**}$ \\ $\sim\lambda^{-(a\mu+b)}$ }
 & \makecell{ PL \\ $\sim\lambda^{-(\frac{1}{2}\mu+1)}$ } 
 & No edge.
 & \makecell{ Frechet   } \\
\hline
\makecell{Heavy tail, \\ $0 < \mu < 2$}
 & \makecell{ (Very) \\ Heavy-Tailed }
 & \makecell{ PL$^{**}$ \\ $\sim\lambda^{-(\frac{1}{2}\mu+1)}$ }
 & \makecell{ PL \\ $\sim\lambda^{-(\frac{1}{2}\mu+1)}$ }  
 & No edge.
 & \makecell{  Frechet } \\
\hline
\end{tabular}%
\end{center}
\caption{Basic MP theory, and the spiked and Heavy-Tailed extensions we use, including known, empirically-observed, and conjectured relations between them.
         Boxes marked ``$^{*}$'' are best described as following ``TW with large finite size corrections'' that are likely Heavy-Tailed~\cite{heavytails2007}, leading to bulk edge statistics and far tail statistics that are indistinguishable.
         Boxes marked ``$^{**}$'' are phenomenological fits, describing large ($2 < \mu < 4$) or small ($0 < \mu < 2$) finite-size corrections on $N\rightarrow\infty$ behavior.
See~\cite{
DPS14,
heavytails2007, disordered2007,
peche_edge,
AAP09,
AGP16,
AT16,
BJ09_TR,
bouchaud2009,
BM97}
for additional details.
}
\label{table:mp_vanilla_spiked_ht}
\end{table}

MP-based RMT is applicable to a wide range of matrices;
but it is \emph{not} in general applicable when matrix elements are strongly-correlated.
Strong correlations appear to be the case for many well-trained, production-quality DNNs.
In statistical physics, it is common to \emph{model} strongly-correlated systems by Heavy-Tailed distributions~\cite{SornetteBook}.
The reason is that these models exhibit, more or less, the same large-scale statistical behavior as natural phenomena in which strong correlations exist~\cite{SornetteBook,bouchaud2009}.
Moreover, recent results from MP/RMT have shown that new Universality classes exist for matrices with elements drawn from certain Heavy-Tailed distributions~\cite{bouchaud2009}. 

We use these Heavy-Tailed extensions of basic MP/RMT to build an operational and phenomenological theory of Regularization in Deep Learning;
and we use these extensions to justify our analysis of both \emph{Self-Regularization} and \emph{Heavy-Tailed Self-Regularization}. % 
Briefly, our theory for simple \emph{Self-Regularization} is insipred by the Spiked-Covariance model of Johnstone~\cite{johnstone2009} and it's interpretation as a form of \emph{Self-Organization} by Sornette~\cite{sornette2002}; and our theory for more sophisticated \emph{Heavy-Tailed Self-Regularization} is inspired by the application of MP/RMT tools in quantitative finance by Bouchuad, Potters, and coworkers~\cite{galluccio1998, bouchaud1999, bouchaud2005, heavytails2007, disordered2007, bouchaud2009, bun2017}, as well as the relation of Heavy-Tailed phenomena more generally to \emph{Self-Organized Criticality} in Nature~\cite{SornetteBook}.
Here, we highlight basic results for this generalized MP theory;
see~\cite{
DPS14,
heavytails2007, disordered2007,
peche_edge,
AAP09,
AGP16,
AT16,
BJ09_TR,
bouchaud2009,
BM97}
in the physics and mathematics literature for additional details.

\noindent\textbf{Universality classes for modeling strongly correlated matrices.}
Consider modeling $\mathbf{W}$ as an $N\times M$ random matrix, with elements drawn from a Heavy-Tailed---e.g., a Pareto or Power Law (PL)---distribution:
\begin{equation}
W_{ij} \sim P(x)\sim\dfrac{1}{x^{1+\mu}},\;\;\mu>0  .
\label{eqn:ht-distribution}
\end{equation}

\noindent
In these cases, if $\mathbf{W}$ is element-wise Heavy-Tailed, %
then the ESD $\rho_{N}(\lambda)$ likewise exhibits Heavy-Tailed properties, either globally for the entire ESD and/or locally at the bulk edge.

Table~\ref{table:mp_vanilla_spiked_ht} summarizes these 
recent results, comparing basic MP theory, the Spiked-Covariance model, %
and Heavy-Tailed extensions of MP theory, including associated Universality classes.
To apply the MP theory, at finite sizes, to matrices with elements drawn from a Heavy-Tailed distribution of the form given in Eqn.~(\ref{eqn:ht-distribution}), 
we have one of the following three %
Universality classes.  
\begin{itemize}
\item
\textbf{(Weakly) Heavy-Tailed},
$4<\mu$:
Here, the ESD $\rho_{N}(\lambda)$ exhibits ``vanilla'' MP behavior in the infinite limit, and the expected mean value of the bulk edge is $\lambda^{+}\sim M^{-2/3}$. 
Unlike standard MP theory, which exhibits TW statistics at the bulk edge, here the edge exhibits PL / Heavy-Tailed fluctuations at finite $N$.
These finite-size effects appear in the edge / tail of the ESD, and they make it hard or impossible to distinguish the edge versus the tail at finite $N$. 
\item
\textbf{(Moderately) Heavy-Tailed},
$2<\mu<4$:
Here, the ESD $\rho_{N}(\lambda)$ is Heavy-Tailed / PL in the infinite limit, approaching 
$\rho(\lambda)\sim\lambda^{-1-\mu/2}$. 
In this regime, 
there is no bulk edge. 
At finite size, the global ESD can be modeled by 
$\rho_{N}(\lambda)\sim\lambda^{-(a\mu+b)}$, for all $\lambda>\lambda_{min}$, but the slope $a$ and intercept $b$ must be fit, as they display 
large finite-size effects.
The maximum eigenvalues follow Frechet (not TW) statistics, with $\lambda_{max}\sim M^{4/\mu-1}(1/Q)^{1-2/\mu}$, and they have large finite-size effects. 
Thus, at any finite $N$, $\rho_{N}(\lambda)$ is Heavy-Tailed, but the tail decays moderately quickly. 
\item
\textbf{(Very) Heavy-Tailed},
$0<\mu<2$: 
Here, the ESD $\rho_{N}(\lambda)$ is Heavy-Tailed / PL for all finite $N$, and as $N\rightarrow\infty$ it converges more quickly to a PL distribution with tails $\rho(\lambda)\sim\lambda^{-1-\mu/2}$. 
In this regime, there is no bulk edge, and the maximum eigenvalues follow Frechet (not TW) statistics. 
Finite-size effects exist, but they are are much smaller here than in the $2<\mu<4$ regime of~$\mu$.  
\end{itemize}

\noindent\textbf{Fitting PL distributions to ESD plots.}
Once we have identified PL distributions visually,
we can fit the ESD to a PL in order to obtain the exponent $\alpha$.
We use the Clauset-Shalizi-Newman (CSN) approach~\cite{CSN09_powerlaw}, as implemented in the python PowerLaw package~\cite{ABP14},%
\footnote{See \url{https://github.com/jeffalstott/powerlaw}.}.
Fitting a PL has many subtleties, most beyond the scope of this paper~\cite{CSN09_powerlaw, GMY04, MPS05, Bau07, KYP11, DC13, ABP14, VC14, HCLT17}.

\noindent\textbf{Identifying the Universality class.}
Given $\alpha$, we identify the corresponding $\mu$ 
and thus which of the three Heavy-Tailed Universality classes ($0<\mu < 2$ or $2 < \mu < 4$ or $4 < \mu$, as described in Table~\ref{table:mp_vanilla_spiked_ht}) is appropriate to describe the system.
The following are particularly important points.
First, observing a Heavy-Tailed ESD may indicate the presence of a scale-free DNN.
This suggests that the underlying DNN is strongly-correlated, and that we need more than just a few separated spikes, plus some random-like bulk structure, to model the DNN and to understand DNN regularization. 
Second, this does not necessarily imply that the matrix elements of $\mathbf{W}_{l}$ form a Heavy-Tailed distribution.
Rather, the Heavy-Tailed distribution arises since we posit it as a model of the strongly correlated, highly non-random matrix $\mathbf{W}_{l}$.
Third, we conjecture that this is more general, and that very well-trained DNNs will exhibit Heavy-Tailed behavior in their ESD for many the weight matrices.

\section{Empirical Results: ESDs for Existing, Pretrained DNNs} 
\label{sxn:empirical_pretrained_models}

In this section, we describe our main empirical results for existing, pretrained DNNs. %
Early on, we observed that small DNNs and large DNNs have very different ESDs.
For smaller models, ESDs tend to fit the MP theory well, with well-understood deviations, e.g., low-rank perturbations.
For larger models, the ESDs $\rho_{N}(\lambda)$ almost never fit the theoretical $\rho_{mp}(\lambda)$, and they frequently have a completely different 
form.
We use RMT to compare and contrast the ESDs of a smaller, older NN and many larger, modern DNNs.
For the small model, we retrain a modern variant of one of the very early and well-known Convolutional Nets---LeNet5. 
For the larger, modern models, we examine selected layers from AlexNet, InceptionV3, and many other models (as distributed with~pyTorch).

\noindent\textbf{Example: LeNet5 (1998).}
LeNet5
is the prototype early model for DNNs~\cite{LBBH98}.
Since LeNet5 is older, we actually recoded and retrained it.
We used Keras 2.0, using $20$ epochs of the AdaDelta optimizer, on the MNIST data set. 
This model has $100.00\%$ training accuracy, and $99.25\%$ test accuracy on the default MNIST split.
We analyze the ESD of the FC1 Layer.
The FC1 matrix $\mathbf{W}_{FC1}$ is a $2450\times500$ matrix, with $Q=4.9$, and thus it yields 500 eigenvalues.

\begin{figure}[t]
\centering
    %\subfigure[Full ESD, FC1, LeNet5]{
    \subfigure[LeNet5, full]{
        \includegraphics[scale=0.26]{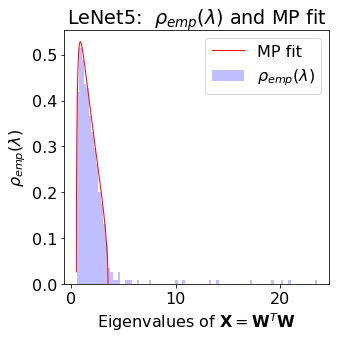}
        \label{fig:lenet5-full}
    }
    %\subfigure[Zoomed-in ESD, FC1, LeNet5]{
    \subfigure[LeNet5, zoomed-in]{
        \includegraphics[scale=0.26]{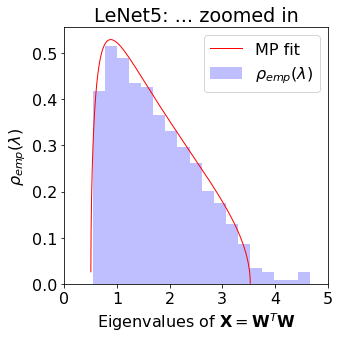}
        \label{fig:lenet5-zoomed}
    }
    %\subfigure[Full ESD, FC2, AlexNet]{
    \subfigure[AlexNet, full]{
        \includegraphics[scale=0.26]{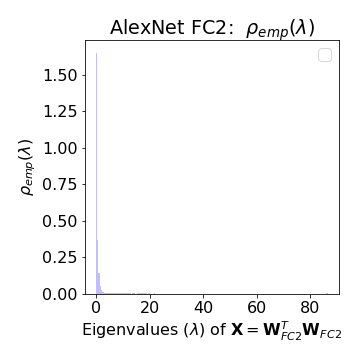} 
        \label{fig:alexnet-fc2-full}
    } 
    %\subfigure[Zoomed-in ESD, FC2, AlexNet]{
    \subfigure[AlexNet, zoomed-in]{
        \includegraphics[scale=0.26]{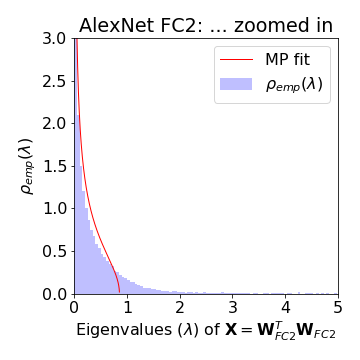} 
        \label{fig:alexnet-fc2-zoom}
    } 
    \caption{Full and zoomed-in ESD for LeNet5 (Layer FC1) and AlexNet (Layer FC2).  
             Overlaid (in red) are fits of the MP distribution (which fit the bulk very well for LeNet5 but \emph{not} well for AlexNet).
    }
   \label{fig:lenet5_and_alexnet_iclr}
\end{figure}

Figures~\ref{fig:lenet5-full} and~\ref{fig:lenet5-zoomed} present
the ESD for FC1 of LeNet5, with Figure~\ref{fig:lenet5-full} showing the full ESD
and Figure~\ref{fig:lenet5-zoomed} 
zoomed-in along the X-axis.
We show (red curve) our fit to the MP distribution $\rho_{emp}(\lambda)$. 
Several things are striking. 
First, the \emph{bulk} of the density $\rho_{emp}(\lambda)$ has a large, MP-like shape for eigenvalues $\lambda < \lambda^{+} \approx 3.5$, and the MP distribution fits this part of the ESD \emph{very} well, including the fact that the ESD just below the best fit $\lambda^{+}$ is concave.
Second, \emph{some eigenvalue mass is bleeding out} from the MP bulk for $\lambda\in[3.5,5]$, although it is quite small.
Third, beyond the MP bulk and this bleeding out region, are several \emph{clear outliers, or spikes}, ranging from $\approx5$ to $\lambda_{max}\lesssim 25$.
Overall, 
the shape of $\rho_{emp}(\lambda)$, the quality of the global bulk fit, and the statistics and crisp shape of the local bulk edge all agree well with 
MP theory augmented with a low-rank~perturbation.

\noindent\textbf{Example: AlexNet (2012).}
AlexNet was the first modern DNN~\cite{KSH12_alexnet}. 
AlexNet resembles a scaled-up version of the LeNet5 architecture; it consists of 5 layers, 2 convolutional, followed by 3 FC layers (the last being a softmax classifier). %
We refer to the last 2 layers before the final softmax as layers FC1 and FC2, respectively.
FC2 has a $4096 \times 1000$ matrix, with $Q=4.096$.

Consider 
AlexNet FC2 
(full in Figures~\ref{fig:alexnet-fc2-full}, and zoomed-in in~\ref{fig:alexnet-fc2-zoom}). 
This ESD differs even more profoundly from standard MP theory. 
Here, we could find no good MP fit.
The best MP fit (in red) does not fit the Bulk part of $\rho_{emp}(\lambda)$ well.  %at all. 
The fit suggests there should be significantly more bulk eigenvalue mass (i.e., larger empirical variance) than actually observed.
In addition, 
the bulk edge is indeterminate by inspection.
It is only defined by the crude fit we present, and any edge statistics obviously do not exhibit TW behavior.
In contrast with MP curves, which are convex near the bulk edge, the entire ESD is concave (nearly) everywhere. 
Here, a PL fit gives good fit $\alpha \approx 2.25$, 
indicating a $\mu \lesssim 3$.
For this layer (and others), 
the shape of $\rho_{emp}(\lambda)$, the quality of the global bulk fit, and the statistics and shape of the local bulk edge are poorly-described by standard MP theory.

\noindent\textbf{Empirical results for other pre-trained DNNs.}
We have also examined the properties of a wide range of other pre-trained models, and we have observed similar Heavy-Tailed properties
to AlexNet
in all of the larger, state-of-the-art DNNs, including VGG16, VGG19, ResNet50, InceptionV3, etc. 
Space constraints prevent a full presentation of these results, but
several observations can be made.
First, all of our fits, except for certain layers in InceptionV3, appear to be in the range $1.5<\alpha \lesssim 3.5$ (where the CSN method is known to perform well).
Second, we also check to see whether PL is the best fit by comparing the distribution to a Truncated Power Law (TPL), as well as an exponential, stretch-exponential, and log normal distributions. 
In all cases, we find either a PL or TPL fits best (with a p-value $\le0.05$), with TPL being more common for smaller values of $\alpha$.
Third, even when taking into account the large finite-size effects in the range $2<\alpha<4$, 
nearly all of the ESDs appear to fall into the $2<\mu<4$ Universality class.

\noindent\textbf{Towards a Theory of Self-Regularization.}
For older and/or smaller models, like LeNet5, the \emph{bulk} of their ESDs $(\rho_{N}(\lambda);\;\lambda\ll\lambda^{+})$ can be well-fit to theoretical MP density $\rho_{mp}(\lambda)$, potentially with 
distinct, outlying \emph{spikes} $(\lambda>\lambda^{+})$. 
This is consistent with the Spiked-Covariance model of Johnstone~\cite{johnstone2009}, a simple perturbative extension of the standard MP theory. %
This is also reminiscent of traditional Tikhonov regularization, in that there is a ``size scale'' $(\lambda^{+})$ separating signal (spikes) from noise (bulk).
This demonstrates that the DNN training process itself engineers a form of implicit \emph{Self-Regularization} into the trained~model.

For large, deep, state-of-the-art DNNs, our observations suggest that there are profound deviations from traditional RMT.
These networks are reminiscent of strongly-correlated disordered-systems that exhibit Heavy-Tailed behavior. 
What is this regularization, and how is it related to our observations of implicit Tikhonov-like regularization on LeNet5? 

To answer this, recall that similar behavior arises in 
strongly-correlated physical systems, where it is known that strongly-correlated systems can be \emph{modeled} by random matrices---with entries drawn from
non-Gaussian Universality classes~\cite{SornetteBook}, e.g., PL or other Heavy-Tailed distributions.
Thus, when we observe that $\rho_{N}(\lambda)$ has Heavy-Tailed properties, we can hypothesize that $\mathbf{W}$ is strongly-correlated,%
\footnote{For DNNs, these correlations arise in the weight matrices during Backprop training (at least when training on data of reasonable-quality).  That is, the weight matrices ``learn'' the correlations in the data.} 
and we can model it with a Heavy-Tailed distribution.
Then, upon closer inspection, we find that the ESDs of large, modern DNNs behave as expected---when using the lens of Heavy-Tailed variants of RMT.
Importantly, unlike the Spiked-Covariance case, which has a scale cut-off $(\lambda^{+})$, in these very strongly Heavy-Tailed cases, correlations appear on every size scale, and we can not find a clean separation between the MP bulk and the spikes.  
These observations demonstrate that modern, state-of-the-art DNNs exhibit a new form of \emph{Heavy-Tailed Self-Regularization}.

\section{5+1 Phases of Regularized Training}
\label{sxn:5plus1phases}

In this section, we develop an operational/phenomenological theory for DNN Self-Regularization.

\noindent\textbf{MP Soft Rank.}
We first define 
the \emph{MP Soft Rank} ($\mathcal{R}_{mp}$), that is designed to capture the ``size scale'' of the noise part of 
$\mathbf{W}_{l}$, relative to the largest eigenvalue of $\mathbf{W}^{T}_{l}\mathbf{W}_{l}$.
Assume that MP theory fits \emph{at least a bulk} of $\rho_{N}({\lambda})$.
Then, we can identify a bulk edge $\lambda^{+}$ and a bulk variance $\sigma^{2}_{bulk}$, and define the \emph{MP Soft Rank} as the ratio of $\lambda^{+}$ and $\lambda_{max}$:
$ %%\begin{equation}
\mathcal{R}_{mp}(\mathbf{W}):=\lambda^{+}/\lambda_{max}  .
$ %%\end{equation}
Clearly, $\mathcal{R}_{mp}\in[0,1]$;
$\mathcal{R}_{mp}=1$ for a purely random matrix; 
and
for a matrix with an ESD with outlying spikes,
$\lambda_{max}>\lambda^{+}$, and $\mathcal{R}_{mp}<1$.
If there is no good MP fit because the entire ESD is well-approximated by a Heavy-Tailed distribution,
then we can define $\lambda^{+}=0$,
in which case $\mathcal{R}_{mp}=0$.

\begin{table}[!htb]
\small
\begin{center}
\begin{tabular}{|p{1in}|c|c|c|c|c|}
\hline

 & \makecell{Operational \\ Definition}
 & \makecell{Informal \\ Description \\ via Eqn.~(\ref{eqn:w_plus_delta_phases}) }
 & \makecell{Edge/tail \\ Fluctuation \\ Comments}
 & \makecell{ Illustration \\ and \\ Description }
 \\
\hline
 \hline
\textsc{Random-like}
 & \makecell{ESD well-fit by MP \\ with appropriate $\lambda^{+}$}
 & \makecell{$\mathbf{W}^{rand}$ random; \\ $\|\Delta^{sig}\|$ zero or small }
 & \makecell{$\lambda_{max} \approx \lambda^+$ is \\ sharp, with \\ TW statistics }
 & \makecell{
Fig.~\ref{fig:phases_of_training-random}
}
 \\
\hline
\textsc{Bleeding-out}
 & \makecell{ESD \textsc{Random-like}, \\ excluding eigenmass \\ just above $\lambda^{+}$}
 & \makecell{$\mathbf{W}$ has eigenmass  at \\ bulk edge as \\ spikes ``pull out''; \\ $\|\Delta^{sig}\|$ medium }
 & \makecell{BPP transition, \\ $\lambda_{max}$ and \\ $\lambda^+$ separate }
 & \makecell{
             Fig.~\ref{fig:phases_of_training-bleedingOut}
             }
 \\
\hline
\textsc{Bulk+Spikes}
 & \makecell{ESD \textsc{Random-like} \\ plus $\ge 1$ spikes \\ well above $\lambda^{+}$}
 & \makecell{$\mathbf{W}^{rand}$ well-separated \\ from low-rank $\Delta^{sig}$; \\ $\|\Delta^{sig}\|$ larger }
 & \makecell{$\lambda^+$ is TW, \\ $\lambda_{max}$ is \\ Gaussian }
 & \makecell{
             Fig.~\ref{fig:phases_of_training-bulkPlusSpike}
             }
 \\
\hline
\textsc{Bulk-decay}
 & \makecell{ ESD less \textsc{Random-like}; \\ Heavy-Tailed eigenmass \\ above $\lambda^{+}$; some spikes }
 & \makecell{Complex $\Delta^{sig}$ with \\ correlations that \\ don't fully enter spike}
 & \makecell{ Edge above $\lambda^+$ \\ is not concave}
 & \makecell{
             Fig.~\ref{fig:phases_of_training-bulkDeterioration}
             }
 \\
\hline
\textsc{Heavy-Tailed}
 & \makecell{ESD better-described \\ by Heavy-Tailed RMT \\ than Gaussian RMT }
 & \makecell{ $\mathbf{W}^{rand}$ is small; \\ $\Delta^{sig}$ is large and \\ strongly-correlated }
 & \makecell{ No good $\lambda^+$; \\ $\lambda_{max} \gg \lambda^+$ }
 & \makecell{
             Fig.~\ref{fig:phases_of_training-heavyTailed}
             }
 \\
\hline
\textsc{Rank-collapse}
 & \makecell{ESD has large-mass \\ spike at $\lambda=0$ }
 & \makecell{ $\mathbf{W}$ very rank-deficient; \\ over-regularization }
 & ---
 & \makecell{
             Fig.~\ref{fig:phases_of_training-singularity}
             }
 \\
\hline
\end{tabular}%
\end{center}
\caption{The 5+1 phases of learning we identified in DNN training.
We observed \textsc{Bulk+Spikes} and \textsc{Heavy-Tailed} in existing trained models (LeNet5 and AlexNet/InceptionV3, respectively; see Section~\ref{sxn:empirical_pretrained_models}); and we exhibited all 5+1 phases in a simple model (MiniAlexNet; see Section~\ref{sxn:main-batch}).
}
\label{table:phases_of_training}
\end{table}

\begin{figure}[!htb]
        \begin{center}
                \subfigure[\textsc{Random-like}.]{
                        \includegraphics[width=0.23\textwidth]{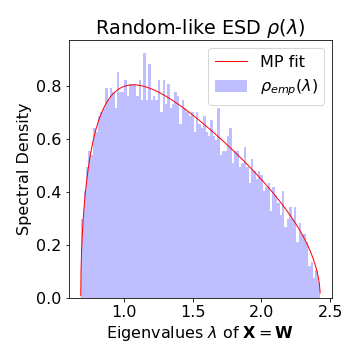} \quad
                        \label{fig:phases_of_training-random}
                }
                \qquad
                \subfigure[\textsc{Bleeding-out}.]{
                        \includegraphics[width=0.23\textwidth]{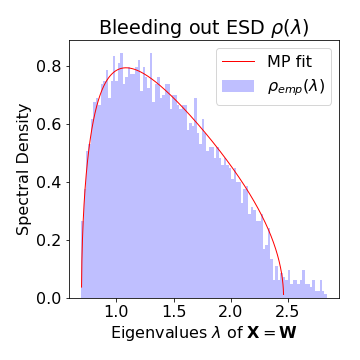} \quad
                        \label{fig:phases_of_training-bleedingOut}
                }
                \qquad
                \subfigure[\textsc{Bulk+Spikes}.]{
                        \includegraphics[width=0.23\textwidth]{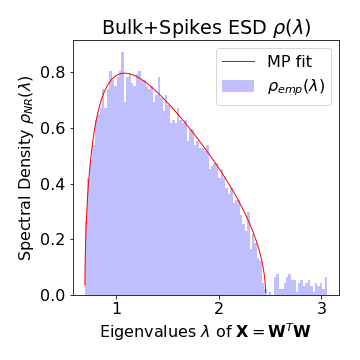} \quad
                        \label{fig:phases_of_training-bulkPlusSpike}
                } \\ 
                \subfigure[\textsc{Bulk-decay}.]{
                        \includegraphics[width=0.23\textwidth]{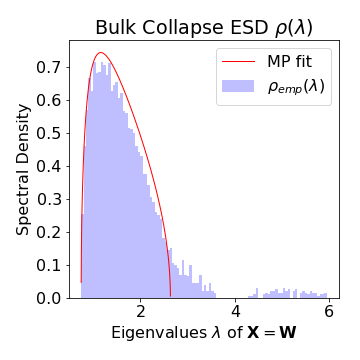} \quad
                        \label{fig:phases_of_training-bulkDeterioration}
                }
                \qquad
                \subfigure[\textsc{Heavy-Tailed}.]{
                        \includegraphics[width=0.23\textwidth]{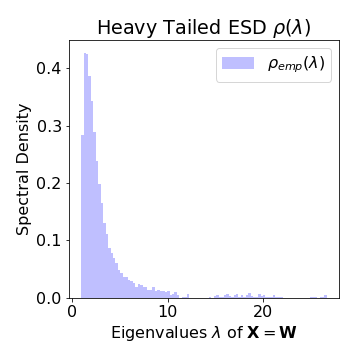} \quad
                        \label{fig:phases_of_training-heavyTailed}
                }
                \qquad
                \subfigure[\textsc{Rank-collapse}.]{
                        \includegraphics[width=0.23\textwidth]{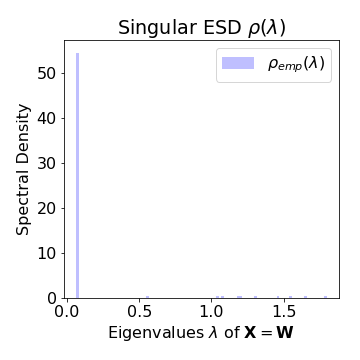} \quad
                        \label{fig:phases_of_training-singularity}
                }
        \end{center}
\caption{%
Taxonomy of trained models.
Starting off with an initial random or \textsc{Random-like} model
(\ref{fig:phases_of_training-random}),
training can lead to a \textsc{Bulk+Spikes} model
(\ref{fig:phases_of_training-bulkPlusSpike}),
with data-dependent spikes on top of a random-like bulk.
Depending on the network size and architecture, properties of training data, etc., additional training can lead to a \textsc{Heavy-Tailed} model
(\ref{fig:phases_of_training-heavyTailed}),
a high-quality model with long-range correlations.
An intermediate \textsc{Bleeding-out} model
(\ref{fig:phases_of_training-bleedingOut}),
where spikes start to pull out from the bulk,
and 
an intermediate \textsc{Bulk-decay} model
(\ref{fig:phases_of_training-bulkDeterioration}), 
where correlations start to degrade the separation between the bulk and spikes, leading to a decay of the bulk,
are also possible.
In extreme cases, a severely over-regularized model
(\ref{fig:phases_of_training-singularity})
is possible.
}
\label{fig:phases_of_training}
\end{figure}

\noindent\textbf{Visual Taxonomy.}
We characterize \emph{implicit Self-Regularization}, both for DNNs during SGD training as well as for \emph{pre-trained} DNNs, as a visual taxonomy of \emph{5+1 Phases of Training} (\textsc{Random-like}, \textsc{Bleeding-out}, \textsc{Bulk+Spikes}, \textsc{Bulk-decay}, \textsc{Heavy-Tailed}, and \textsc{Rank-collapse}).
See Table~\ref{table:phases_of_training} for a summary.
The 5+1 phases can be ordered, with each successive phase corresponding to a smaller Stable Rank / MP Soft Rank and to progressively more Self-Regularization than previous phases.
Figure~\ref{fig:phases_of_training} depicts typical ESDs for each phase, with the MP fits (in red). 
Earlier phases of training correspond to the final state of older and/or smaller models like LeNet5 and MLP3.
Later phases correspond to the final state of more modern models like AlexNet, Inception, etc.
While we can describe this in terms of SGD training, this taxonomy 
allows us to compare different architectures and/or amounts of regularization in a trained---or even pre-trained---DNN. 

Each phase is visually distinct, and each has a natural interpretation in terms of RMT.
One consideration is the \emph{global properties of the ESD}: how well all or part of the ESD is fit by an MP distriution, for some value of $\lambda^+$, or how well all or part of the ESD is fit by a Heavy-Tailed or PL distribution, for some value of a PL parameter.
A second consideration is \emph{local properties of the ESD}: the form of fluctuations, in particular around the edge $\lambda^+$ or around the largest eigenvalue $\lambda_{max}$.
For example, the shape of the ESD near to and immediately above $\lambda^+$ is very different in Figure~\ref{fig:phases_of_training-random} and Figure~\ref{fig:phases_of_training-bulkPlusSpike} (where there is a crisp edge) versus Figure~\ref{fig:phases_of_training-bleedingOut} (where the ESD is concave) versus Figure~\ref{fig:phases_of_training-bulkDeterioration} (where the ESD is convex).

\noindent\textbf{Theory of Each Phase.}
RMT provides more than simple visual insights, and we can use RMT to differentiate between the \emph{5+1 Phases of Training} using simple models that qualitatively describe the shape of each ESD.
We model the weight matrices $\mathbf{W}$ as ``noise plus signal,'' where the ``noise'' is modeled by a random matrix $\mathbf{W}^{rand}$, with entries drawn from the Gaussian Universality class (well-described by traditional MP theory) and the ``signal'' is a (small or 
large) correction~$\Delta^{sig}$:
\begin{equation}
\mathbf{W}\simeq\mathbf{W}^{rand}+\Delta^{sig}   .
\label{eqn:w_plus_delta_phases}
\end{equation}
Table~\ref{table:phases_of_training} summarizes the theoretical model for each phase.
Each model uses RMT to describe 
the global shape of $\rho_{N}(\lambda)$,
the local shape of the fluctuations at the bulk edge, and
the statistics and information in the outlying spikes,
including possible Heavy-Tailed behaviors.

In the first phase (\textsc{Random-like}), the ESD is well-described by traditional MP theory, in which a random matrix has entries drawn from the Gaussian Universality class.
In the next phases (\textsc{Bleeding-out}, \textsc{Bulk+Spikes}), and/or for small networks such as LetNet5, $\Delta$ is a relatively-small perturbative correction to $\mathbf{W}^{rand}$, and vanilla MP theory (as reviewed in Section~\ref{sxb:review-RMT-MP}) can be applied, as least to the bulk of the ESD.
In these phases, we  will \emph{model} the $\mathbf{W}^{rand}$ matrix by a vanilla $\mathbf{W}_{mp}$ matrix (for appropriate parameters), and the MP Soft Rank is relatively large $(\mathcal{R}_{mp}(\mathbf{W})\gg 0)$.
In the \textsc{Bulk+Spikes} phase, the model resembles a Spiked-Covariance model, and the Self-Regularization resembles Tikhonov regularization.  

In later phases (\textsc{Bulk-decay}, \textsc{Heavy-Tailed}), and/or for modern DNNs such as AlexNet and InceptionV3, $\Delta$ becomes more complex and increasingly dominates over $\mathbf{W}^{rand}$.
For these more strongly-correlated phases, $\mathbf{W}^{rand}$ is relatively much weaker, and the MP Soft Rank decreases. 
Vanilla MP theory is not appropriate, and instead the Self-Regularization becomes Heavy-Tailed.
We will treat the noise term $\mathbf{W}^{rand}$ as small, and we will \emph{model} the properties of $\Delta$ with Heavy-Tailed extensions of vanilla MP theory (as reviewed in Section~\ref{sxb:review-RMT-extensions}) to Heavy-Tailed non-Gaussian universality classes that are more appropriate to model strongly-correlated systems.
In these phases, the strongly-correlated model is still regularized, but in a very non-traditional way.
The final phase, the \textsc{Rank-collapse} phase, is a degenerate case that is a prediction of the~theory.

\section{Empirical Results: Detailed Analysis on Smaller Models}
\label{sxn:main}

To validate and illustrate 
our theory, we analyzed MiniAlexNet,%
\footnote{\url{https://github.com/deepmind/sonnet/blob/master/sonnet/python/modules/nets/alexnet.py}}
a simpler version of AlexNet, similar to the smaller models used in \cite{Understanding16_TR}, scaled down to prevent overtraining, and trained on CIFAR10. 
Space constraints prevent a full presentation of these results, but we mention a few key results here.
The basic architecture consists of two 2D Convolutional layers, each with Max Pooling and Batch Normalization, giving 6 initial layers; it then has two Fully Connected (FC), or Dense, layers with ReLU activations; and it then has a final FC layer added, with $10$ nodes and softmax activation. 
$\mathbf{W}_{FC1}$ is a $4096\times 384$ matrix ($Q \approx 10.67$);
$\mathbf{W}_{FC2}$ is a $384\times 192$ matrix ($Q = 2$); and 
$\mathbf{W}_{FC3}$ is a $192\times 10$ matrix.    
All models are trained using Keras 2.x, with TensorFlow as a backend. 
We use SGD with momentum, with a learning rate of $0.01$, a momentum parameter of $0.9$, and a baseline batch size of $32$; and we train up to $100$ epochs.
We save the weight matrices at the end of every epoch, and we 
analyze
the empirical properties of the $\mathbf{W}_{FC1}$ and $\mathbf{W}_{FC2}$ matrices.

\begin{figure}[!htb]
   \centering
   \subfigure[Epoch 0]{
      \includegraphics[scale=0.29]{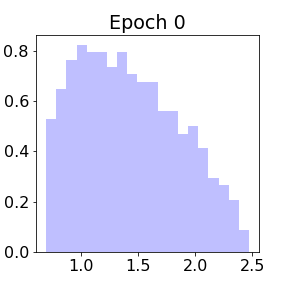}
   }
   \subfigure[Epoch 4]{
      \includegraphics[scale=0.29]{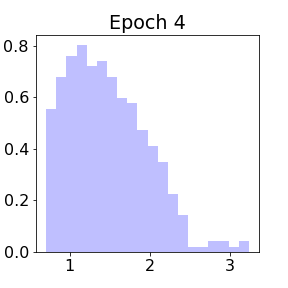}
   }
   \subfigure[Epoch 8]{
      \includegraphics[scale=0.29]{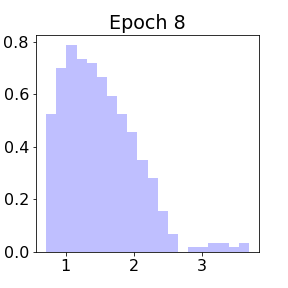}
   }
   \subfigure[Epoch 12]{
      \includegraphics[scale=0.29]{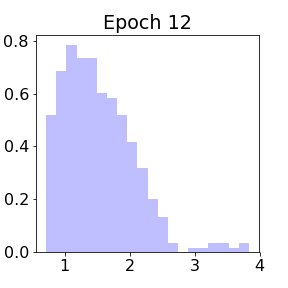}
   }
   \caption{Baseline ESD for Layer FC1 of MiniAlexNet, during training.
}
   \label{fig:w7every5}
\end{figure}

For each layer, the matrix Entropy ($\mathcal{S}(\mathbf{W})$) gradually lowers; and the Stable Rank ($\mathcal{R}_{s}(\mathbf{W})$) shrinks.  
These decreases parallel the increase in training/test accuracies, and both 
metrics level off as the training/test accuracies do.
These changes are seen in the ESD, e.g., see Figure~\ref{fig:w7every5}.
For layer FC1,
the initial weight matrix $\mathbf{W}^{0}$ looks very much like an MP distribution (with $Q \approx 10.67$),
consistent with a \textsc{Random-like} phase.
Within a very few epochs, however, eigenvalue mass shifts to larger values, and the ESD looks like the \textsc{Bulk+Spikes} phase.
Once the Spike(s) appear(s), substantial changes are hard to see 
visually, 
but minor changes do continue in the ESD.
Most notably, $\lambda^{max}$ increases from roughly $3.0$ to roughly $4.0$ during training, indicating further Self-Regularization, even within the \textsc{Bulk+Spikes} phase.
Here, spike eigenvectors tend to be more localized than bulk eigenvectors.
If explicit regularization (e.g., $L_2$ norm weight regularization or Dropout) is added, then we observe a greater decrease in the complexity metrics (Entropies and Stable Ranks), consistent with expectations, and this is casued by the eigenvalues in the spike being pulled to much larger values in the ESD.
We also observe that eigenvector localization tends to be more prominent, presumably since explicit regularization can make spikes more well-separated from the bulk.

\section{Explaining the Generalization Gap by Exhibiting the Phases}
\label{sxn:main-batch}

In this section, we demonstrate that we can exhibit all five of the main phases of learning by changing a single knob of the learning process. %
We consider the batch size 
since it is not traditionally considered a regularization parameter and due to its its implications for the generalization~gap.  %% phenomenon.

The \emph{Generalization Gap} refers to the peculiar phenomena that DNNs generalize significantly less well when trained with larger mini-batches (on the order of $10^{3}-10^{4}$)~\cite{leCun98, HHS17_TR, KMNST16_TR, one_hour17_TR}. 
Practically, this is of interest since smaller batch sizes makes training large DNNs on modern GPUs much less efficient. 
Theoretically, this is of interest since it contradicts simplistic stochastic optimization theory for convex problems.
Thus, 
there is interest in the question: what is the mechanism responsible for the drop in generalization in models trained with SGD methods in the large-batch~regime?

To address this question, we consider here using different batch sizes in the DNN training algorithm.
We trained the MiniAlexNet model, just as in Section~\ref{sxn:main}, except with batch sizes ranging from moderately large to very small ($b\in\{500,250,100,50,32,16,8,4,2\}$).

\begin{figure}[!htb]
    \centering
    \subfigure[Layer FC1.]{
        \includegraphics[width=3.3cm]{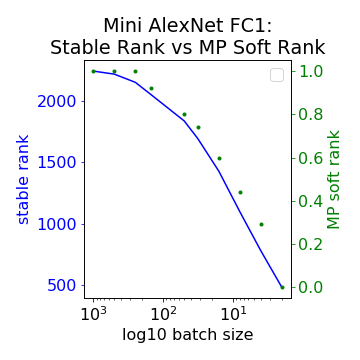} 
        \label{fig:mini-alexnet-softrank-per-batch-fc1}
    }
    \qquad
    \subfigure[Layer FC2.]{
        \includegraphics[width=3.3cm]{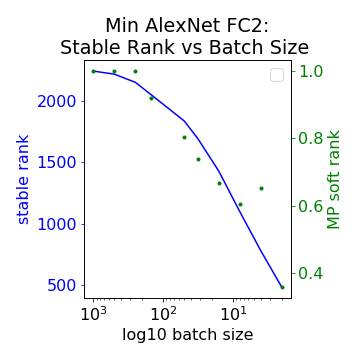} 
        \label{fig:mini-alexnet-softrank-per-batch-fc2}
    }
    \qquad
    \subfigure[Training, Test Accuracies.]{
        \includegraphics[width=3.3cm]{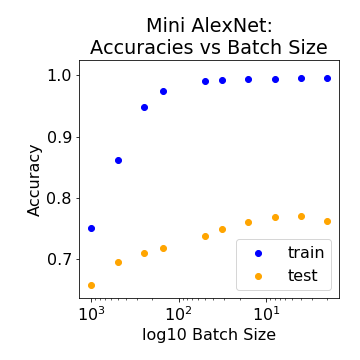}
        \label{fig:mini-alexnet-traintest-per-batch}
   }
    \caption{Varying Batch Size.
             Stable Rank and MP Softrank for FC1 (\ref{fig:mini-alexnet-softrank-per-batch-fc1}) and FC2 (\ref{fig:mini-alexnet-softrank-per-batch-fc2}); and Training and Test Accuracies (\ref{fig:mini-alexnet-traintest-per-batch}) versus Batch Size for MiniAlexNet.
}
    \label{fig:mini-alexnet-softrank-traintest-per-batch}
\end{figure}

\noindent\textbf{Stable Rank, MP Soft Rank, and Training/Test Performance.}
Figure~\ref{fig:mini-alexnet-softrank-traintest-per-batch} shows the Stable Rank and MP Softrank for FC1 (\ref{fig:mini-alexnet-softrank-per-batch-fc1}) and FC2 (\ref{fig:mini-alexnet-softrank-per-batch-fc2}) as well as the Training and Test Accuracies (\ref{fig:mini-alexnet-traintest-per-batch}) as a function of Batch Size.
The MP Soft Rank ($\mathcal{R}_{mp}$) and the Stable Rank ($\mathcal{R}_{s}$) both track each other, and both systematically \emph{decrease} with decreasing batch size, as the test accuracy \emph{increases}. 
In addition, both the training and test accuracy decrease for larger values of $b$: training accuracy is roughly flat until batch size $b \approx 100$, and then it begins to decrease; and test accuracy actually increases for extremely small $b$, and then it gradually decreases as $b$ increases.

\begin{figure}[t]
    \centering
 %   \subfigure[Batch Size 1000.]{
 %       \includegraphics[width=2.7cm]{img/mini-alexnet-fc1-bs1000.png} 
   %     \label{fig:lmini-alexnet-fc1-bs1000}
  %  }
 %   \qquad
    \subfigure[Batch Size 500.]{
        \includegraphics[width=2.7cm]{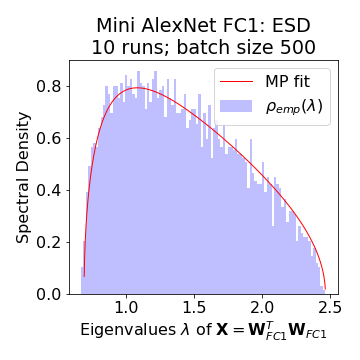} 
        \label{fig:mini-alexnet-fc1-bs500}
    }
     \qquad
    \subfigure[Batch Size 250.]{
        \includegraphics[width=2.7cm]{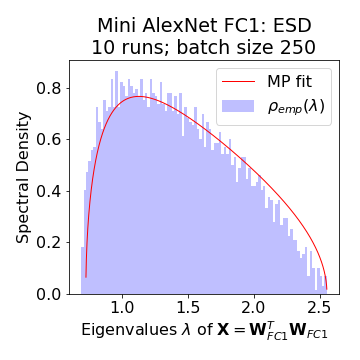} 
        \label{fig:mini-alexnet-fc1-bs250}
    }
     \qquad
    \subfigure[Batch Size 100.]{
        \includegraphics[width=2.7cm]{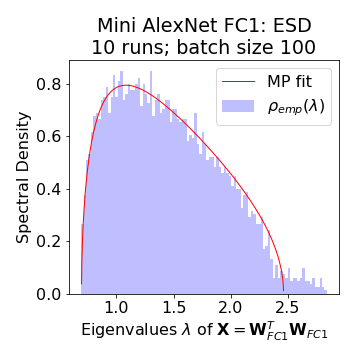} 
        \label{fig:mini-alexnet-fc1-bs100}
    }
         \qquad
    \subfigure[Batch Size 32.]{
        \includegraphics[width=2.7cm]{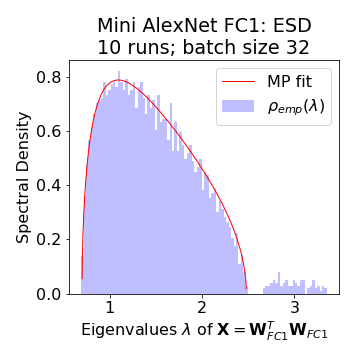} 
        \label{fig:mini-alexnet-fc1-bs32}
    }
             \qquad
    \subfigure[Batch Size 16.]{
        \includegraphics[width=2.7cm]{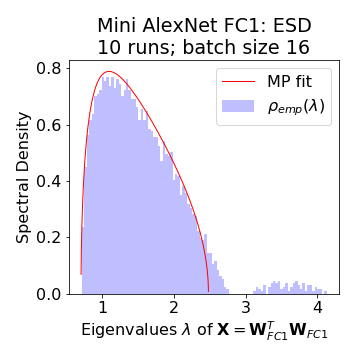} 
        \label{fig:mini-alexnet-fc1-bs16}
    }
         \qquad
    \subfigure[Batch Size 8.]{
        \includegraphics[width=2.7cm]{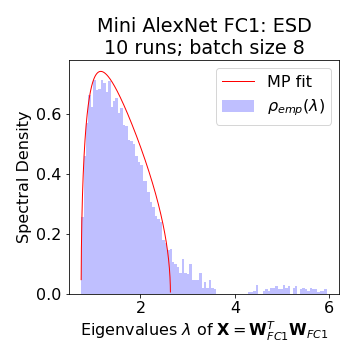} 
        \label{fig:mini-alexnet-fc1-bs8}
    }
                 \qquad
    \subfigure[Batch Size 4.]{
        \includegraphics[width=2.7cm]{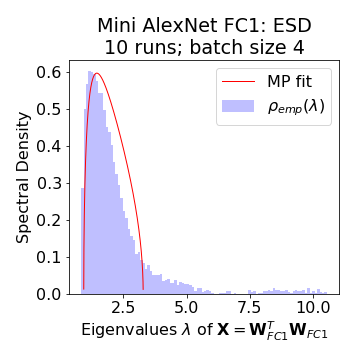} 
        \label{fig:mini-alexnet-fc1-bs14}
    }
         \qquad
    \subfigure[Batch Size 2.]{
        \includegraphics[width=2.7cm]{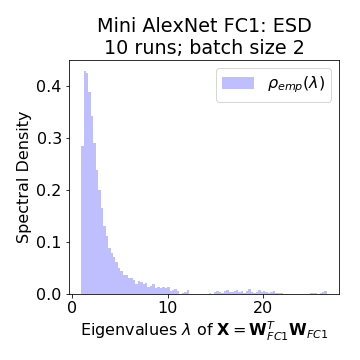} 
        \label{fig:mini-alexnet-fc1-bs2}
    }
    \caption{Varying Batch Size.
             ESD for Layer FC1 of MiniAlexNet, with MP fit (in red), for an ensemble of 10 runs, for Batch Size ranging from $500$ down to $2$.
             %%\michael{MM to do: what are other training knobs, or is this the baseline setup, except for varying $b$.}
             %%\charles{DO you want this both in the text and the caption?  or just text ?}\michael{Both is okay, as it's important enough.}
             Smaller batch size leads to more implicitly self-regularized models.
             We exhibit all 5 of the main phases of training by varying only the batch size.
     }
  \label{fig:alexnet-fc1-batches}
\end{figure}

\noindent\textbf{ESDs: Comparisons with RMT.}
Figure~\ref{fig:alexnet-fc1-batches} shows the final ensemble ESD for each value of $b$ for Layer FC1.  %% and FC2, respectively, of MiniAlexNet.
We see systematic changes in the ESD as batch size $b$ decreases.
At batch size $b=250$ (and larger), the ESD resembles a pure MP distribution with no outliers/spikes; it is \textsc{Random-like}.
As $b$ decreases, there starts to appear an outlier region.
For $b=100$, the outlier region resembles \textsc{Bleeding-out}.
For $b=32$, these eigenvectors become well-separated from the bulk, and the ESD resembles \textsc{Bulk+Spikes}. 
As batch size continues to decrease, the spikes grow larger and spread out more (observe the 
scale of the X-axis), and the ESD exhibits \textsc{Bulk-decay}.
Finally, at 
$b=2$, extra mass from the main part of the ESD plot almost touches the spike, and the curvature of the ESD changes, consistent with \textsc{Heavy-Tailed}.
In addition, as $b$ decreases, some of the extreme eigenvectors associated with eigenvalues that are not in the bulk tend to be more localized.

\noindent\textbf{Implications for the generalization gap.}
Our results here (both that training/test accuracies decrease for larger batch sizes and that smaller batch sizes lead to more well-regularized models) demonstrate that the generalization gap phenomenon arises since, for smaller values of the batch size $b$, the DNN training process itself implicitly leads to stronger Self-Regularization.
(This Self-Regularization can be either the more traditional Tikhonov-like regularization or the Heavy-Tailed Self-Regularization corresponding to strongly-correlated models.)
That is, training with smaller batch sizes implicitly leads to more well-regularized models, and it is this regularization that leads to improved results.
The obvious mechanism is that, by training with smaller batches, the DNN training process is able to ``squeeze out'' more and more finer-scale correlations from the data, leading to more strongly-correlated models.
Large batches, involving averages over many more data points, simply fail to see this very fine-scale structure, and thus they are less able to construct strongly-correlated models characteristic of the \textsc{Heavy-Tailed} phase.

\section{Discussion and Conclusion}
\label{sxn:discussion}

Clearly, our theory opens the door to address numerous very practical questions.  %%, among them the following.
One of the most obvious is whether our RMT-based theory is applicable to other types of layers such as convolutional layers.
Initial results suggest yes, but the situation is more complex than the relatively simple picture we have described here.
These and related directions are promising avenues to explore.

\bibliographystyle{unsrt}
%\bibliographystyle{plain}

%{\small
\bibliography{dnns,gen_gap}
%}

\end{document}